\documentclass[runningheads]{llncs}
\usepackage{eccv}
\usepackage{algorithm}
\usepackage{algpseudocode}
\usepackage{eccvabbrv}
\usepackage{graphicx}
\usepackage{booktabs}
\usepackage[accsupp]{axessibility} 
\usepackage{hyperref}
\begin{document}
\title{\textit{Sketch \& Paint}: Stroke-by-Stroke Evolution of  Visual Artworks} 
\titlerunning{\textit{Sketch \& Paint}}
\author{Jeripothula Prudviraj, \and
Vikram Jamwal}

\authorrunning{J.~Prudviraj  and J.~Vikram}
\institute{TCS Research, INDIA \\
\email{\{prudviraj.jeripothula,vikram.jamwal\}@tcs.com}}

\maketitle

\begin{figure}[h!]
    \centering
    \includegraphics[width= \textwidth]{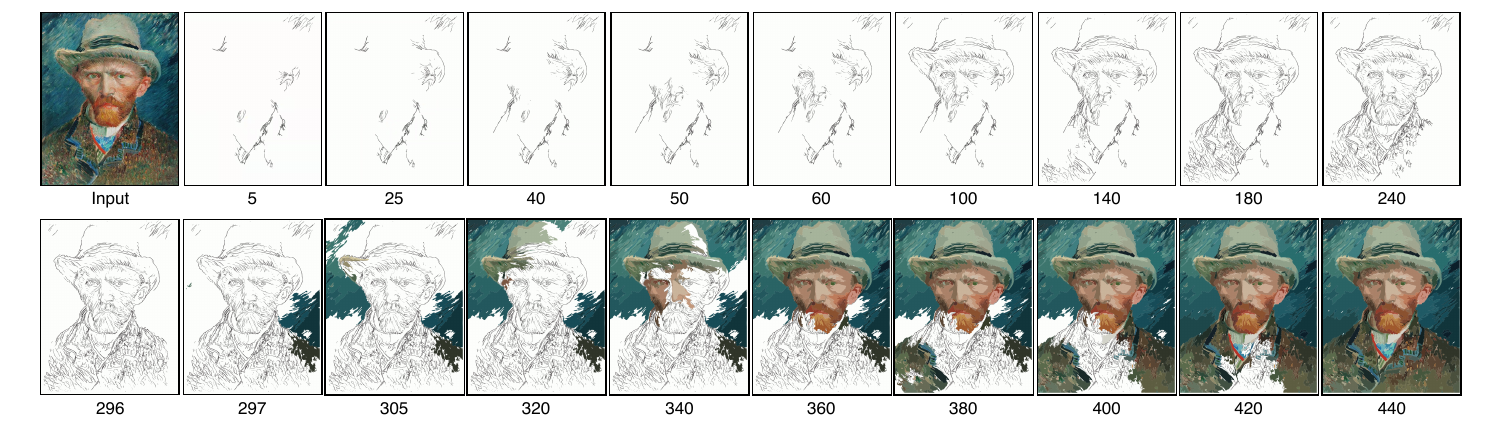}
    \caption{Illustration of constructed stroke sequence for input visual art}
    \label{fig:Constructed}
\end{figure}

\begingroup
\let\clearpage\relax
\begin{abstract}
Understanding the stroke-based evolution of visual artworks is useful for advancing artwork learning, appreciation, and interactive display. While the stroke sequence of renowned artworks remains largely unknown, formulating this sequence for near-natural image drawing processes can significantly enhance our understanding of artistic techniques. This paper introduces a novel method for approximating artwork stroke evolution through a proximity-based clustering mechanism. We first convert pixel images into vector images via parametric curves and then explore the clustering approach to determine the sequence order of extracted strokes. Our proposed algorithm demonstrates the potential to infer stroke sequences in unknown artworks. We evaluate the performance of our method using WikiArt data and qualitatively demonstrate the plausible stroke sequences. Additionally, we demonstrate the robustness of our approach to handle a wide variety of input image types such as line art, face sketches, paintings, and photographic images. By exploring stroke extraction and sequence construction, we aim to improve our understanding of the intricacies of the art development techniques and the step-by-step reconstruction process behind visual artworks, thereby enriching our understanding of the creative journey from the initial sketch to the final artwork. 
\end{abstract}
\section{Introduction}
Visual art is an essential part of humanity. It allows us to explore, express, and communicate ideas, emotions, perspectives, and experiences. Visual arts encompass various mediums such as drawing, painting, sculpture, and photography. In the realm of drawings and paintings, we encounter a wide range of styles, themes, and artistic movements that reflect the creativity and cultural diversity of human existence. Understanding the complex landscape of art development is important for art education and appreciation.

Digital technology can help bridge the gap between traditional art forms and modern accessibility \cite{brown2020routledge, carrozzino2010beyond}. For example, high-resolution digital reconstructions for artworks allow museums to create virtual exhibits with interactive features such as zooming into specific painting parts or viewing various stroke forms of artwork, thereby enhancing visitor understanding and engagement \cite{carvajal2020virtual, carrozzino2010beyond}.  Similarly, understanding the \textit{process of constructing a painting} is extremely educative.  Investigating the extraction of strokes from famous artworks and constructing a stroke order can provide significant insights into the creative process and foster a deeper understanding of artistic craftsmanship and innovation.  Moreover, detailed dynamic digital reconstruction of artworks also serves as a valuable educational resource \cite{sylaiou2017exploring, lockee2014visual}, enabling students and researchers to closely study historical paintings, understand the used techniques, and explore the artist’s process. It can be surmised that by studying and emulating the techniques of master artists through digital reconstructions, students can develop their skills in painting and drawing, gaining insights into fundamental aspects such as brush control, color mixing, and composition. 

Interactive models that help understand the drawing process, therefore, can prove to be extremely useful. However, building methods that mimic the pragmatic drawing process is extremely challenging. It is hard to define the semantics of each piece of art, the geometries of sketch/painting, the number and the order of strokes, the individual stroke attributes like length, color, shape, \& texture, and the overall evolution of an artwork.  The fundamental challenges in building the interactive drawing model are: (i) How do we represent and extract stroke-level information for complex art (ii) What is the plausible way to construct the sequence order of strokes that mimic a drawing process?

Some works have explored the explication of the drawing process in the recent past. Fu \textit{et al} \cite{fu2011animated} introduce an algorithm designed to animate pre-drawn line drawings by determining stroke order, though the method struggles with accuracy over complex sketches. Subsequently, Sketch-RNN \cite{ha2017neural} learns the construction of stroke sequences of hand-drawn sketches by training an ML model on thousands of human-drawn images. However, its learning is based on the available labeled stroked data and it does not learn directly from the sketches. Furthermore, the method does not scale for complex sketches involving shading and textures. Recently, Tong \textit{et al.} \cite{tong2021sketch} show advancement with an image-to-pencil translation method that produces 
sketches and demonstrates the drawing process. Additionally, several works \cite{vinker2023clipascene, vinker2022clipasso} have investigated new levels of abstraction in object sketching through geometric and semantic simplifications. Despite these advancements, accurately recreating the intricate details of complex artworks remains a significant challenge. Furthermore, neural painting techniques \cite{liu2021paint, huang2019learning, song2024processpainter} employing reinforcement learning have attempted to generate stroke sequences for non-photo-realistic image recreation. Nevertheless, these methods struggle with the high computational demands of deep reinforcement learning and lack an inherent sequence order while generating strokes.

With these observations, we explore the extraction of stroke sequences from artworks, aiming to construct a step-by-step process, that an artist might use for the evolution of an artwork from an initial sketch to the final complete painting (E.g., as shown in Figure \ref{fig:Constructed}). We leverage Scalable Vector Graphics (SVG) 
\cite {zhang2023towards,liu2009beyond,carlier2020deepsvg,mateja2023animatesvg} to extract stroke-level data through inverse graphics as in \cite{das2020beziersketch, romaszko2017vision,kulkarni2015deep,rodriguez2023starvector}. We choose to represent the input image in terms of parameterized curves such as B\'{e}zier curves due to their simple structure and encoding ability of complex and finer details of the input image.

As discussed, the fundamental requirements of the interactive generation model are stroke-by-stroke data of input and the sequence order in which the evolution takes place, as if the art is drawn by a human. To achieve this, in our work,  we explore perceptual grouping inspired by the theory of Gestalt laws \cite{wagemans2012century} to reason out the sequence order for unlabeled stroke data and mimic the pragmatic drawing process for sketching and painting.
 
Our work differentiates from many of the present methods such as animated drawing \cite{fu2011animated}, Sketch-RNN \cite{ha2017neural}, CLIPasso \cite{vinker2022clipasso}, or paint transformer \cite{liu2021paint} in that we suggest an integrated mechanism for the evolution of artwork from both \textit{sketch and paint} perspective and that we aim at recreating the original artwork. The constructed stroke sequence not only enables building interactive sketching and painting models but can also be useful for downstream tasks like art completion, manipulation, generation, and retrieval. 

The contributions of our work can be summarized as: 

\begin{itemize}
    \item We present a simple yet effective procedural algorithm to construct the stroke sequencing order for stroke evolution. The robustness of the proposed method enables us to handle large numbers of strokes, thereby facilitating the examination of complex images.
    \item To the best of our knowledge, we are the first to address the stroke-by-stroke evolution of complex artworks and natural images. 
    \item In comparison to other methods that limit stroke order to sketches, our work, termed \textit {sketch \& paint}, presents the stroke-by-stroke sequencing for both sketching and color painting.
    \item Further, we demonstrate the generalizability of the proposed method on various forms of visual data like sketches, clip art, and natural images. We validate the efficacy of the proposed method using samples extracted from publicly available datasets like WikiArt, VectroFlow, and FS2K-SDE.
\end{itemize}

\newpage
\section{Related work}
A few works in the past have explored stroke-based methods for interactive generations. The work of Ivan Sutherland \cite{sutherland1964sketch} is the first to investigate interactive interfaces for freehand drawing. A pencil rendering technique is presented in \cite{sousa1999observational} through observation models to simulate artists and illustrators. Chen \textit {et al.} \cite{chen2004example} investigated a method for portrait drawing based on composite sketching. An input image is divided into several layers in \cite {li2003feature} to render the intensity of each stacked layer. Though these methods can generate sketches, they fail to offer the drawing process and only produce the final result. 

Humans create art through a stroke-by-stroke mechanism rather than pixel-wise operations. Towards this, Fu \textit{et al.} \cite{fu2011animated} presented an algorithm that leverages human-drawn line drawings in order to extract stroke order and then animate the sketch. In particular, they proposed a method to estimate drawing order from static line drawings by applying conventional principles of drawing order. 
This approach utilizes Hamiltonian graph minimization and an energy function to determine stroke order. While the method is efficient in smaller search spaces, it becomes more complex as the number of strokes increases. The most computationally demanding step involves finding Hamiltonian paths on k-nn graphs, with running times ranging from a few seconds to 2 to 5 minutes. This depends on the value of k, the number of significant lines, and the structure of the k-nn graphs. The method is limited to line art images with clearly defined lines or curves, excluding those with shading, texture, or complex geometric sketches that are difficult to distinguish. The inputs are also assumed to be relatively clean and free of hatching strokes. Also, they establish that the order of detail strokes is less crucial, allowing them to use a simpler strategy rather than the computationally intensive one required for significant lines. 
In comparison, our proposed work operates on complex images with a large number of strokes.

Liu et al. addressed the problem of simulating the process of observational drawing, focusing on how people draw lines when sketching a given 3D model. They presented a multiphase drawing framework in which drawing actions are ordered by phases: posture phase, primitive phase, contour phase, and details phase. The lines within these phases are organized at three levels: phase-by-phase, part-by-part, and finally, stroke-by-stroke. To measure the information gained between previously drawn strokes and the target drawing as ground truth, they build a graph similar to \cite{fu2011animated} adopting the greedy Prim's minimum spanning tree algorithm. However, this method cannot be extended to complex images with a large number of strokes similar to \cite{fu2011animated}. Further, an RNN-based method, Sketch-RNN \cite{ha2017neural}, is explored on a human-drawn image to construct stroke-based drawings of common objects. It mainly utilizes the pen-state information of the digitally drawn sketch to learn stroke sequences. However, this approach uses only simple hand-drawn objects (QuickDraw) with few strokes and does not scale to real paintings. Moreover, QuickDraw is prone to sampling noise due to highly correlated temporal sequences and suffers from limited capacity as presented in \cite{das2020beziersketch}. 

Zheng \cite{zheng2018strokenet} presented a StrokeNet that can generate a sequence of strokes toward Chinese character writing. 
However, the generated sequence and their strokes are far from human writing. To progress in AI-assisted creative sketching, Songwei \textit{et al.} \cite{ge2020creative} introduced Creative Birds and Creative Creatures datasets, where they proposed a part-based GAN to predict suggestions for partial sketches by generating novel part compositions. 
Though this method generates compositional parts, it does not go with human creative flow construction. Further, Yonggang \textit{et al.} \cite {qi2021sketchlattice} introduced an alternative sketch representation based on the lattice structure over a 2D plane towards the sketch manipulation task. All these methods investigate the simple single object categories and do not consider complex natural or art data.

To tackle natural images, various works \cite{vinker2022clipasso,vinker2023clipascene} presented a photo-to-sketch method to convert scene to sketch by different levels of abstraction. CLIPasso \cite{vinker2022clipasso} presents a photo-to-sketch method to convert a single object image to a sketch by different levels of abstraction. Here, sketches are derived from a set of B\'{e}zier curves and the number of strokes defines the level of abstraction. 
CLIPascene \cite{vinker2023clipascene} is an extension of CLIPasso, where it extends a single object category to a scene. Though these methods produce vector curves, they are very sparse and not suitable for faithful sketching due to limited details. Tong et al. \cite{tong2021sketch} introduced the drawing process for image-to-pencil sketches by drawing one stroke at a time.  At first, they established a parameter-controlled pencil stroke generation mechanism based on the pixel-scale statistical results of some real pencil drawings and then exploited a framework to guide stroke arrangement on the canvas. Here, they determine stroke using central pixel gray value, line width, and line length. And, they use an Edge Tangent Flow (ETF) vector field to guide the direction of the stroke. However, the ETFs do not have inherent sequence order and do not enable the natural drawing process. Further, the representation of pencil lines is one form of sketching and does not support a wide variety of complex data.

Few works \cite {liu2021paint, huang2019learning} explored Reinforcement Learning (RL) based mechanisms where the objective is to predict a set of B\'{e}zier curves through rendering to minimize the difference between the rendered image and the target. Even though these methods can generate high-quality paintings, they generate random curves on canvas and do not hold any inherent sequence. Specifically, Hung et al. \cite{huang2019learning} employed the Deep Deterministic Policy Gradient (DDPG) algorithm to train a neural agent for oil painting. 
However, their Deep Reinforcement Learning (DRL) approach faces limitations due to its requirement for a large number of parameters, constraining the network input size to 128x128 images. This constraint limits the generation of fine-grained details. In contrast, our algorithm does not impose any restrictions on input image size and is capable of generating high-quality, detailed results.

\newpage
\section{Methodology}
In this work, we formulate the problem of stroke sequence generation in visual art, where our goal is to build a model that can produce a pragmatic drawing process for input art. Given an input image $I$, our objective is to construct a stroke sequence $Stroke\_seq$ = $(seq_0, seq_1, seq_2, \ldots seq_N$) such that each stroke ($seq_n$), when put in a sequence, aligns and constructs the scene semantically, i.e., $Stroke\_seq (I) \approx I$. 
Here, the semantics of the stroke sequence implies the semblance of a pragmatic drawing process. Towards this,  we present an unsupervised stroke sequence generation method in an open-world setting. Fig \ref{fig:Outline} presents the outline of the proposed method, where we decompose input art and sketch into a set of strokes and compose stroke sequencing to give the semblance of a pragmatic drawing process. 

\begin{figure}[!t]
    \centering
    \includegraphics[width=\textwidth]{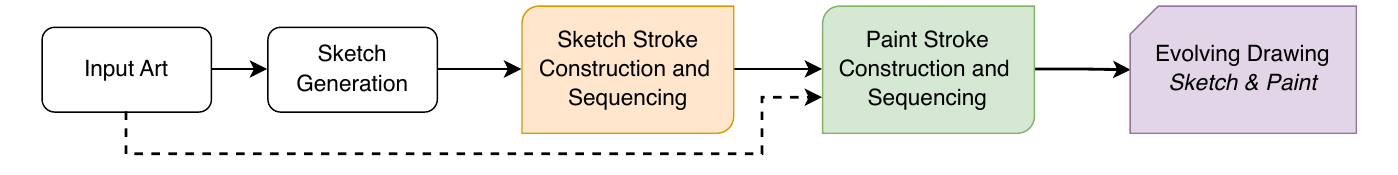}
    \caption{Outline of the proposed method for sketch to paint}
    \label{fig:Outline}
\end{figure}

Given an input image, we first process it through a sketch generator that distills the input painting into a line drawing that effectively extracts the underlying sketch. Specifically, our sketch generator leverages the state-of-the-art line drawing generator \cite{chan2022learning} that builds on depth information and CLIP features. 

\begin{equation}
     S = SketchGenerator (I) 
\end{equation}

Here, $SketchGenerator$ extracts line drawings by probing the geometrics and semantics of an input image using a pre-trained model \cite{chan2022learning}. 
Once we have sketch and original art, we feed these inputs into individual streams of sketch stroke construction and sequencing (Section \ref{sscs}) and paint stroke construction and sequencing (Section \ref{pscs}) to make a unified global sequencing.

\subsection{Sketch Stroke Construction and Sequencing} \label{sscs}
Upon receiving a sketch ($S$), the Sketch Stroke Construction and Sequencing (SSCS) module orchestrates the sequence of strokes via sketch stroke construction and sketch stroke sequence generator, as illustrated in Figure \ref{fig:Sketch_seq}. 
\begin{figure}[ht!]
    \centering
    \includegraphics[width=\textwidth]{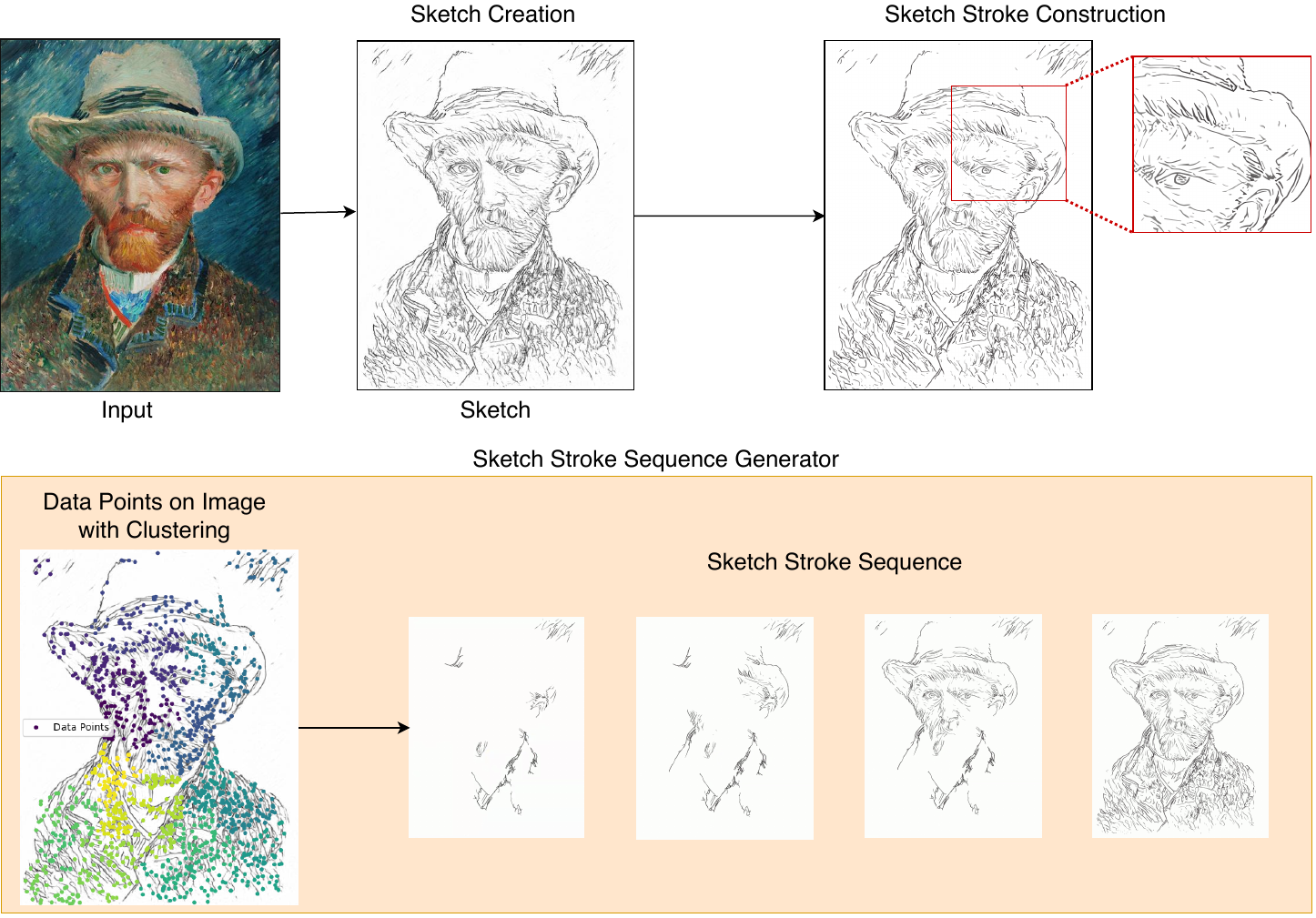}
    \caption{Framework for sketch stroke construction and sequencing}
    \label{fig:Sketch_seq}
\end{figure}

\subsubsection{Sketch Stroke Construction}
Given an input sketch $S$ of a scene, a raster image with a set of pixels, the stroke construction module produces a set of strokes ($S\_Strokes_N$) based on vector curves. In the past, various vector curves have been investigated for stroke design \cite {zhang2023towards,liu2009beyond,carlier2020deepsvg,chan2022learning}. 
In this work, we mainly employ the stroke representation in terms of Line, Quadratic B\'{e}zier Curve (QBC), Cubic B\'{e}zier Curve (CBC), Circular Arc (CA), and Elliptical Arc (EA). Here, the position of control points determines the shape of the curves. The number of control points that define the shape of line, QBC, CBC, CA, AND EA 
are $2$, $3$, $4$, $3$, and $3$, respectively. 
Formally, the control points along with color parameter are defined as:  
\begin{align*}
&[x_0, y_0, x_1, y_1, {\#color}] \longrightarrow (line), \\
&[x_0, y_0, x_1, y_1, x_2, y_2, {\#color}] \longrightarrow (QBC), \\
&[x_0, y_0, x_1, y_1, x_2, y_2, x_3, y_3, {\#color}] \longrightarrow (CBC), \\
&[x_0, y_0, x_1, y_1, x_2, y_2, {\#color}] \longrightarrow (CA), and \\ &[x_0, y_0, x_1, y_1, x_2, y_2, {\#color}]\longrightarrow  (EA). 
\end{align*}

Here, $x_n, y_n$ denote the control points of curves and $\#color$ indicates the associated grey scale curve intensity. The stroke construction module can be stated as: 
\[ S\_Strokes_N = StrokeConstruct(S), \hspace{4mm} \text{where} \hspace{1mm} S\_Strokes_N=\{s_0, s_1, s_2, \ldots , s_N\} \]

Here, $S$ denotes sketch and the $StrokeConstruct$ converts the pixel input into a set of geometric curves in terms of line, QBC, CBC, CA, and EA. This can be achieved through any vectorizing method, such as \cite{vectwebsite}.

\subsubsection{Stroke Sequence Generator}
Upon obtaining the vector stroke set of the input sketch $(S\_Strokes_N)$, we first organize the strokes into coherent groups based on proximity. In the context of perceptual grouping, in Gestalt theory\cite{li2018universal}, \textit{similarity} and \textit{proximity} are the most influential human factors. Various works have explored these factors in the context of image segmentation \cite{chen2017deeplab,wang2017unsupervised,ren2003learning} and sketch segmentation \cite{sun2012free,qi2013sketching,schneider2016example}. 

Since we are only interested in stroke sequencing, we drop the similarity measure and probe only the proximity measure for vector curves to build clusters. Specifically, we explore the Hierarchical Clustering method \cite{zhao2005hierarchical} by calculating pairwise distance on coordinates of the starting control points to generate stroke clusters. 
This notion of proximity-grouping helps us to segment the input sketch into semantically smaller groups. 
The stroke clusters of the input image can be obtained as
\[S\_Stroke\_clusters = Hierarchical\_clustering (S\_Strokes_N, dist_{prox})\] 
Here, $dist_{prox} = 0$ makes each vector as an individual group, and $dist_{prox} = \infty$ forms all vectors as a single group.

\subsubsection{Intra-Cluster Sequencing}With the proximity based hierarchical clustering, each stroke cluster $S\_Stroke\_clusters_i$ within $S\_Stroke\_clusters$ establishes an ordered sequence for the strokes $(s_0, s_1,$ $\ldots, s_n)$ contained in that cluster and thus supports the inherent \textit{intra-cluster} sequencing. Specifically, we exploit Ward’s-based linkage matrix from hierarchical clustering to construct an ordered sequence of strokes within each cluster by minimizing variance and implicitly suggesting proximity-based sequencing.

\subsubsection{Inter-Cluster Sequencing}To further regulate the sequencing order of retrieved clusters, we employ the optimization method on generated clusters. Particularly, we select the center coordinates of each cluster as reference points and adopt the formulation of the classical ``\textit{Travelling Salesman Problem''} (TSP). Formally, given a set of $M$ stroke clusters $S\_Stroke\_Clusters$ = $\{C_1, C_2,$ $\ldots, C_M\}$, 
where each cluster \(C_i\) is represented by its centroid coordinates (\(P_{ij}=(p_i, p_j)\)), 
the objective is to come up with the shortest possible trip that visits each stroke cluster exactly once and returns to the starting cluster. This notion helps to achieve \textit{inter-cluster} sequencing over ordered strokes.  

\[
S\_Stroke\_seq = TSP(S\_Stroke\_Clusters_M (P_{ij}))
\]

Here, the decision variables and objective function remain unaltered, i.e., it uses the binary decision variable to determine the tour between the clusters and the objective function is ${Minimize} \sum_{i=1}^{M} \sum_{j=1, j \neq i}^{M} c_{ij} \cdot b_{ij}$
where \(c_{ij}\) represents the distance between the centroids of stroke clusters \(i\) and \(j\). 
And, \(b_{ij}\) is a binary decision variable indicating whether there is a direct tour between stroke clusters \(C_i\) and \(C_j\). 

\subsection{Paint Stroke Construction and Sequencing} \label{pscs}
Similar to Sketch Stroke Construction and Sequencing (SSCS) as presented above, we formulate RGB stroke sequencing for input art towards colouring the constructed sketch, referred as Paint Stroke Construction and Sequencing (PSCS). 
The layout of the PSCS is depicted in Figure \ref{fig:Paint_seq}. Given an RGB input, our paint stroke construction module harvests the set of coloured vector strokes ($P\_Strokes_T$). 

\begin{figure}[!t]
    \centering
    \includegraphics[width=\textwidth]{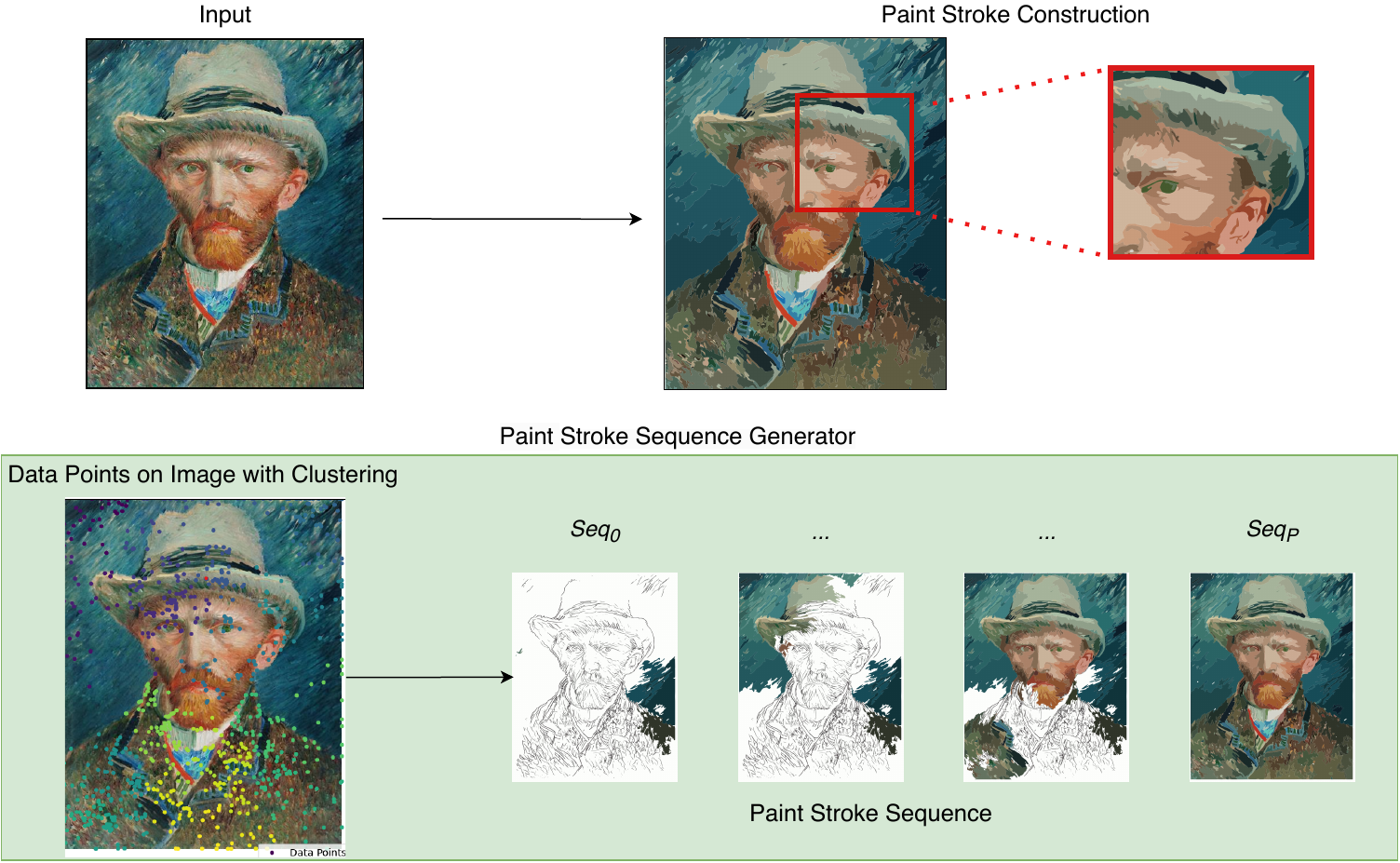}
    \caption{Framework for paint stroke construction and sequencing}
    \label{fig:Paint_seq}
\end{figure} 

\[ P\_Strokes = StrokeConstruct(X) \hspace{4mm} \text{where} \hspace{1mm} P\_Strokes_T={s_0, s_1, s_2, \ldots , s_T.} \]

Here, $X$ denotes input art and the $StrokeConstruct$ produces the RGB geometric curves in terms of line, QBC, CBC, CA, and EA. 
The fundamental difference between sketch strokes and paint strokes lies in their intensity values. Sketch strokes are usually gray-scale strokes, consisting varying shades of gray. 
Whereas, paint strokes utilize RGB color information to capture color appearance. 

As in sketch stroke sequence generator, we employ Hierarchical Clustering method on coordinates of the starting control points of RGB vector curves to articulate semantic segments. 
Further, TSP based optimization method is incorporated on extracted clusters to determine sequencing order. 
\[P\_Stroke\_clusters = Hierarchical\_clustering (P\_Strokes_T, dist_{prox}).\] 
\[
P\_Stroke\_seq = TSP(P\_Stroke\_Clusters_V (Q_{ij}))
\]
Considering that users commonly sketch before applying color, we stream SSCS before integrating it to PSCS, after computing their respective sequence order. 

\begin{equation}
    Stroke\_seq_{\text{Global}} = (S\_Stroke\_seq + P\_Stroke\_seq)
\end{equation}

Here, $Stroke\_seq_{Global}$ portrays the stroke-by-stroke evolution of artworks from initial sketch to final execution.

We summarize our method for \textit{Sketch \& Paint} as Algorithm \ref{alg:cap} below.

\begin{algorithm}
\caption{Sketch \& Paint Stroke Construction and Sequence Generation}\label{alg:cap}
\begin{algorithmic}
\Require Input image $I$
\Ensure Final stroke sequence $Stroke\_seq_{Global}$
\State Read input image $I$
\State Generate Sketch ($S$) from $I$ by generating line drawing of an input image 
    \State \hspace{\algorithmicindent} $Sketch \leftarrow SketchGenerator(I)$
    \State Extract set of strokes via converting pixel sketch into geometric curves
    \State \hspace{\algorithmicindent}$Strokes_{Sketch}$: $S\_Strokes_N \leftarrow StrokeConstruct(S)$
    
    \State Employ proximity based hierarchical clustering on $S\_Strokes_N$
    \State \hspace{\algorithmicindent} $S\_Stroke\_clusters_M \leftarrow \text{Hierarchical\_clustering}(S\_Strokes_N, dist_{\text{proximity}})$
    \State Devise TSP on centroids of stroke clusters
    \State \hspace{\algorithmicindent} $S\_Stroke\_seq = TSP(S\_Stroke\_Clusters_M (P_{ij}))$
\If{painting == TRUE (for an RGB image ($X$))}
        \State Extract set of strokes via converting pixel image into geometric curves
        \State \hspace{\algorithmicindent}$Strokes_{RGB}$: $P\_Strokes_T \leftarrow StrokeConstruct(X)$
        \State Employ proximity based hierarchical clustering on $P\_Strokes_T$
    \State \hspace{\algorithmicindent}$P\_Stroke\_clusters_V \leftarrow \text{Hierarchical\_clustering}(P\_Strokes_T, dist_{\text{proximity}})$
    \State Devise TSP on centroids of stroke clusters
    \State \hspace{\algorithmicindent} $P\_Stroke\_seq = TSP(P\_Stroke\_Clusters_V (Q_{ij}))$
    \EndIf
    \State  $Stroke\_seq_{Global} \leftarrow$ $S\_Stroke\_seq$ + $P\_Stroke\_seq$
    
    \Return $Stroke\_seq_{Global}$
\end{algorithmic}
\end{algorithm}

\section{Experiments}
In this section, we present the data collection and pre-processing, implementation details, and experimental results of the proposed approach. 

\subsection{Data Collection and Implementation Details}
\subsubsection{Dataset}
To the best of our knowledge, there are no existing datasets that demonstrate the stroke-by-stroke evolution of visual art from the initial sketch to the final painting. The closest available dataset for ordered stroke sequences is QuickDraw \cite{ha2017neural}, which consists of simple hand-drawn vector sketches of common objects, created in an online game where players are to draw specific objects within 20 seconds. However, these sketches do not align with our objective of generating the evolution of complex real-world artworks.

For our purposes, we curate a sample dataset from WikiArt \cite{saleh2015large} that includes a diverse range of artworks by renowned artists. In particular, we sampled 500 artworks from various artists to evaluate the effectiveness of the proposed method. Additionally, to investigate the proposed method in diverse settings, we harvested 90 sketch images and 70 RGB images, which include line art, face sketches, and natural images. We randomly sampled face sketches from FS2K-SDE Dataset \cite{dai2023sketch}, line art sketches from \cite{lineartweb}, and natural images from \cite{tong2021sketch}. 

\subsubsection{Implementation details}
To obtain sketches from paintings and natural images, we leverage the line drawing method \cite{chan2022learning} that trained on sampled COCO dataset using CLIP features. Further, to attain a vector image of the input sketch or image, we convert pixel image into vector curves through SVG conversion via the vectorizing tool \cite{vectwebsite}. We impose no restrictions on the dimensions of image inputs or the number of strokes within each image. And, the proximity distance is treated as a hyper-parameter. Here, the number of clusters and the number of strokes per cluster are determined based on this proximity distance. We found that setting proximity distance to approximately 
$\max(Input_{width}, Input_{height})/8$ yields compact clusters that provide a good balance between the number of clusters and the number of strokes per cluster.

\begin{figure}[!t]
    \centering
    \includegraphics[width=\textwidth]{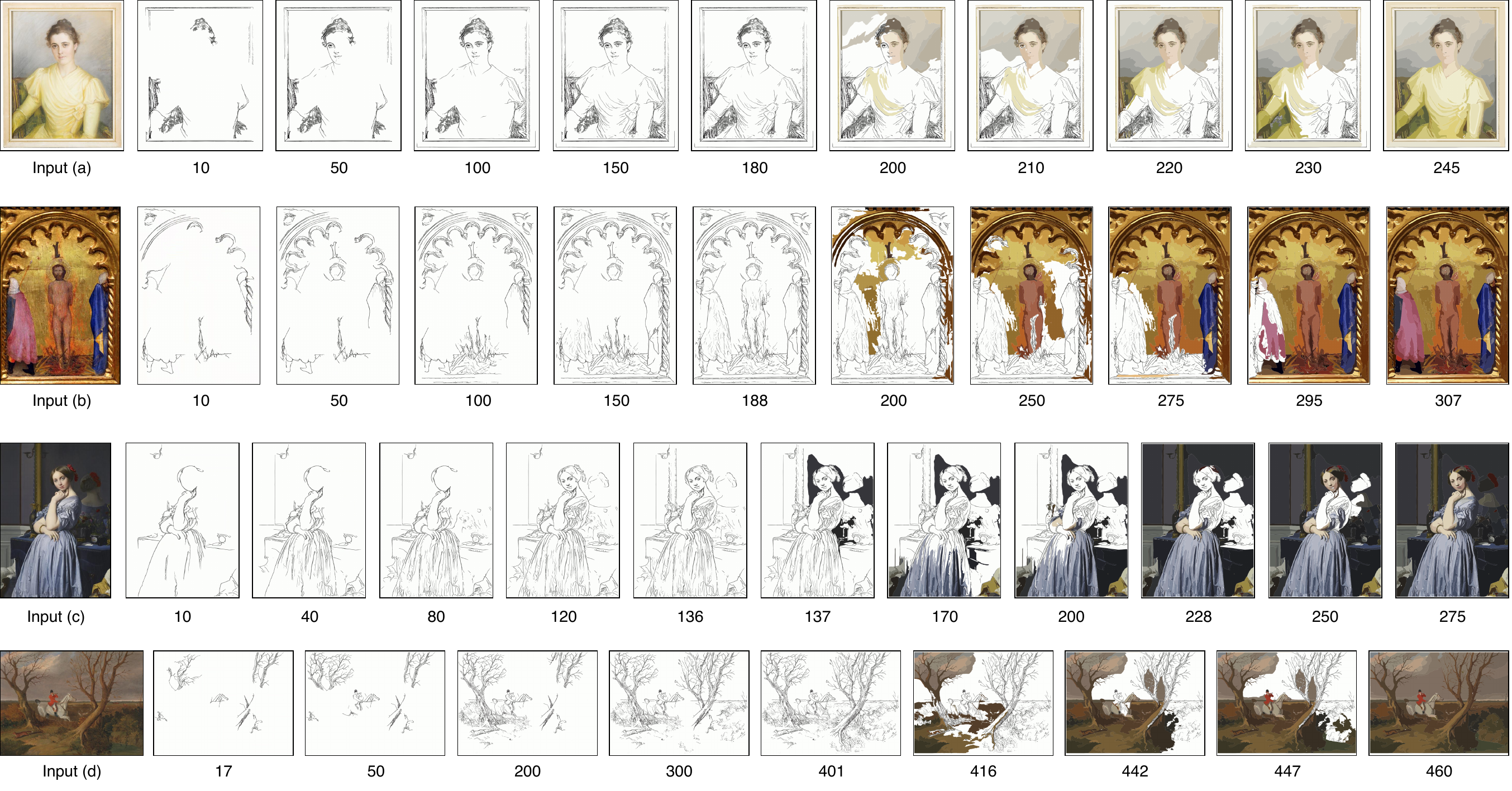}
    \caption{Sketch \& Paint stroke evolution sequences on WikiArt samples.}
    \label{fig:wikiart_results}
\end{figure}

\subsection{Results}
We extensively test our algorithm on inputs with varying degrees of complexity and structure. Since there are no formal quantitative measures to gauge the valid stroke sequence evolution on input images, we principally assess the effectiveness of the proposed model qualitatively.

\subsubsection{Results on WikiArt}
 Figure \ref{fig:wikiart_results} shows some examples of stroke-by-stroke ordering on the WikiArt dataset. From this figure, we can observe that the proposed algorithm successfully composes stroke sequence evolution from sketch to paint. Additionally, it effectively handles images with varied resolutions, intricate details, numerous strokes, and diverse color palettes.

\begin{figure}[!t]

    \centering
    \includegraphics[width=\textwidth]{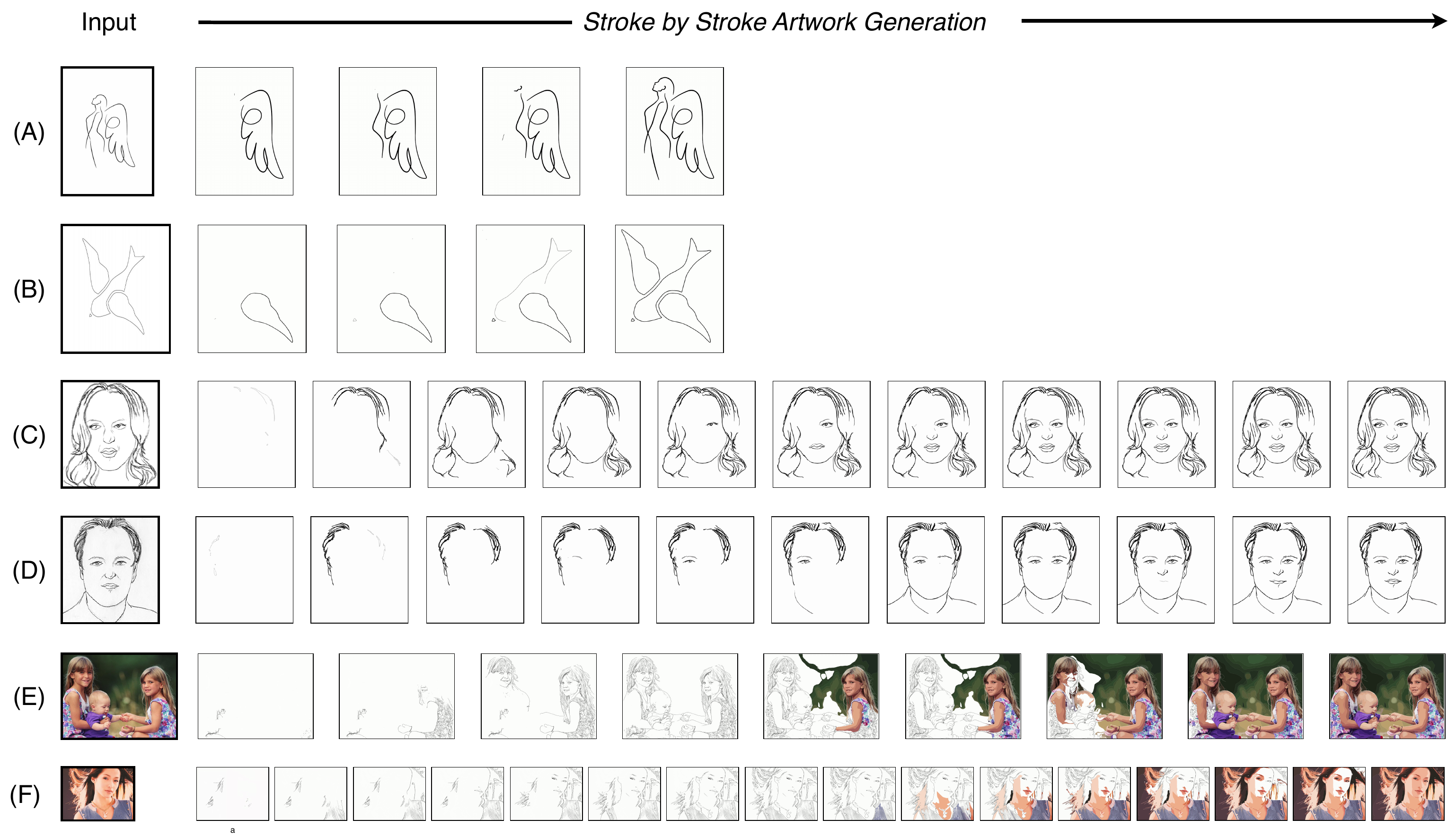}   
    \caption{Demonstration of stroke evolution on various other input data types such as (A-B) Simple line art \cite{lineartweb} (C-D) Face sketches sampled from FS2K-SDE \cite{dai2023sketch}, (E-F) Natural images from \cite{tong2021sketch}.}
    \label{fig:aaai_results}
\end{figure}

To evaluate the robustness of the proposed method, we further apply it to other forms of data such as line art, face sketches, and natural images. Figure \ref{fig:aaai_results} presents sampled sequences from these diverse input images. The results demonstrate that the predicted stroke sequence order closely mirrors a pragmatic drawing process, regardless of the input type. Specifically, the algorithm produces plausible drawing sequences for less complex images like line art (Figure \ref{fig:aaai_results}, A-B) and face sketches (Figure \ref{fig:aaai_results}, C-D). These predicted sequences follow a logical and intuitive order, closely mirroring the natural progression an artist is likely to adopt. As seen in Figure \ref{fig:aaai_results} (E-F), our method can also effectively interpret natural images that are complex in terms of resolution, detail, and stroke count. From all the results presented above, we can infer that our algorithm can comprehend a variety of input images and produce a pragmatic drawing process. Dynamic examples of drawing evolution, from sketch to painting across various inputs, can be viewed at \cite{youtube_video}.

\subsubsection{Comparison with other methods}
In this section, we analyze the effectiveness of our algorithm relative to other state-of-the-art methods. Figure \ref{fig:comparison} illustrates the comparative evolution of our method against prominent techniques \cite{tong2021sketch, liu2021paint}. 
For a fair assessment, we qualitatively compare only our image-to-sketch translation with VectorFlow’s \cite{tong2021sketch} image-to-pencil translation, and only our colored paint stroke sequence with Paint Transformer’s \cite{liu2021paint} paint sequence. 
In other words, to ensure fair comparability, we omit color sequencing for the VectorFlow comparison and sketch sequencing for the Paint Transformer comparison. From Figure \ref{fig:comparison}, we can infer that our proposed method can provide systematic sequencing rather than projecting strokes in random order as in \cite{liu2021paint,tong2021sketch}.

\begin{figure}[!t]
    \centering
    \includegraphics[width=\textwidth]{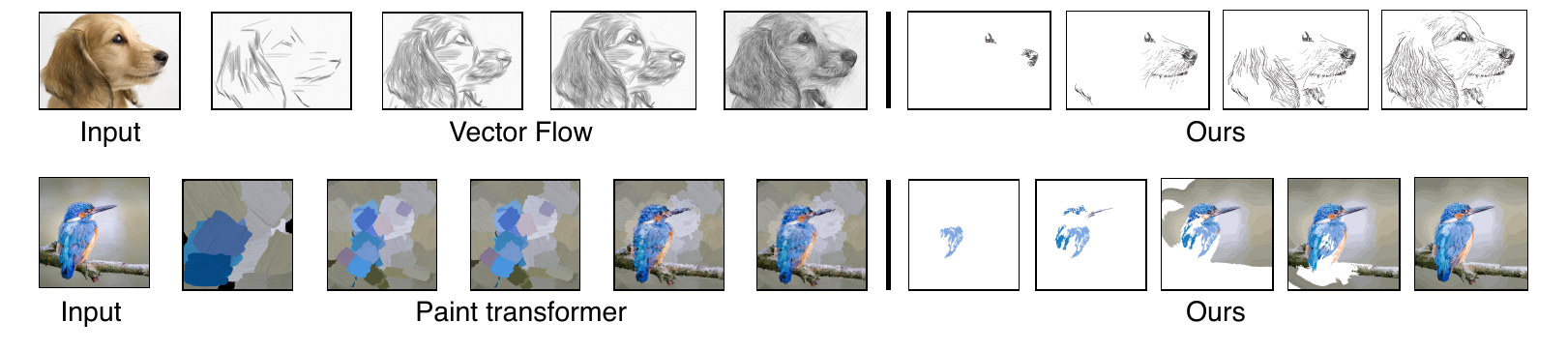}
    \caption{Qualitative comparison of our method over Vector Flow \cite{tong2021sketch} and Paint Transformer \cite{liu2021paint} (We only include the relevant corresponding portions of generated sequence from our method).} 
    \label{fig:comparison}
\end{figure}

\begin{figure}[!t]
    \centering
    \includegraphics[width=\textwidth]{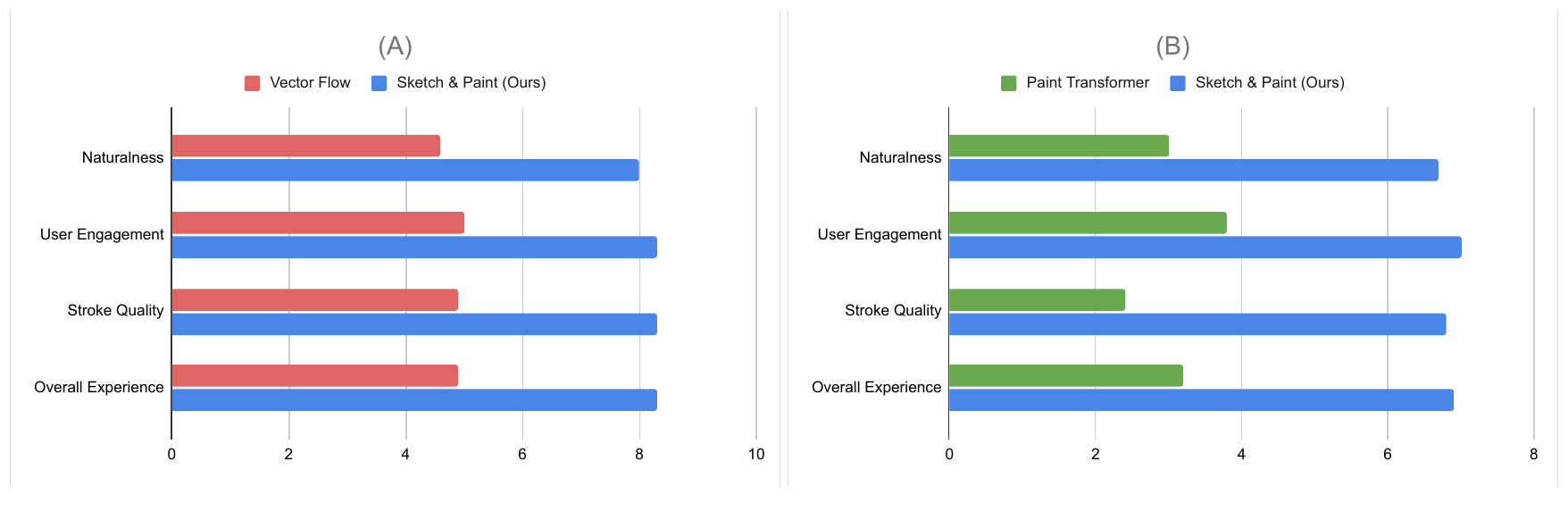}
    \caption{User-survey results for quality evaluation of our method, Paint \& Sketch, in comparison with  Vector Flow (A) \cite{tong2021sketch} and Paint Transformer (B). \cite{liu2021paint} }
    \label{fig:user_survey}
\end{figure}

Additionally, we conducted an initial user study with limited participants (5), each evaluating 5 image generations with VectorFlow \cite{tong2021sketch} and 3 image generations with PaintTransformer \cite{liu2021paint} along with corresponding generations through our method. Participants rated the systems on naturalness, user engagement, stroke quality, and overall experience, with scores ranging from 1 to 10. The survey asked the participants about how accurately the system emulated a pragmatic drawing process, how engaged they felt during the interaction, what the quality of stroke texture and tone was, and what their overall satisfaction was. The bar chart depicting the average user study results is shown in Figure \ref{fig:user_survey}. These results demonstrate that our method provides a more engaging and satisfying user experience than other related methods.

\subsubsection{Limitations} The proposed approach has some known limitations: (1) The method depends upon the quality of line drawing for sketch generation. Hence, it may fail to extract contour lines when the image is dominated by black-intensity regions. (2) There is no mechanism that identifies and learns from the generated sequences that are better aligned to human drawing processes. (3) The paint strokes are coarse and typically not in the form of strokes from any particular drawing medium such as painting brushes.
\section{Discussion and Conclusion}
In this work, we addressed the challenge of constructing stroke sequences for unlabeled artworks - from initial sketch strokes to final painting strokes. Specifically, we formulate a method for ordering stroke sequences through clustering and optimizing vector curves to facilitate a pragmatic drawing process. Our approach significantly advances the understanding of artwork evolution by generating stroke sequences for complex artworks. We validated the effectiveness of our algorithm across various data forms, demonstrating its capability to manage diverse inputs and produce coherent drawing processes. 

Our approach, at the moment, is agnostic to the semantic content of the artwork. The alignment performance of the existing algorithm might further improve if we introduce region-based grouping (for example, by leveraging models such as  Segment Anything Models (SAM)) as artists tend to complete one semantic region before moving on to another. Additionally, we might benefit from the datasets that capture the artist's stroke sequences on the drawings. Alternatively, one can explore using art annotators to establish stroke sequences for a limited number of artworks. Implementing human feedback mechanisms or preference-based modeling could align the algorithm more closely with human drawing styles. We believe that these enhancements would not only deepen our comprehension of the drawing process but also open up possibilities for practical applications, including next-stroke generation, sketch completion, and artwork retrieval. Moreover, this type of work provides a foundation for enriching art education and appreciation by offering insights into the sequential evolution of artworks.

\endgroup

\bibliographystyle{splncs04}
\bibliography{main}
\end{document}